\documentclass{article} 
\usepackage{iclr2017_conference,times}
\usepackage{hyperref}
\usepackage{url}

\usepackage{algorithm}
\usepackage{algorithmic}
\usepackage{amsmath}
\usepackage{amsfonts}
\usepackage{graphics}
\usepackage{epsfig}
\usepackage{subfigure}
\usepackage{float}
\usepackage{multirow}
\usepackage{appendix}
\usepackage{color}
\usepackage{times}
\usepackage{helvet}
\usepackage{courier}
\usepackage{multirow}
\usepackage{bm}
\usepackage[american]{babel}

\iclrfinalcopy
\title{
Loss-aware Binarization of Deep Networks
}

\author{Lu Hou,\;
Quanming Yao,\;
James T. Kwok 
\\
	Department of Computer Science and Engineering\\
	Hong Kong University of Science and Technology\\
	Clear Water Bay, Hong Kong \\
	\texttt{\{lhouab,qyaoaa,jamesk\}@cse.ust.hk} \\
}

\newtheorem{remark}{Remark}[section]
\newtheorem{theorem}{Theorem}[section]
\newtheorem{prop}{Proposition}[section]

\def \h{{\bf h}}
\def \u{{\bf u}}
\def \v{{\bf v}}
\def \w{{\bf w}}
\def \x{{\bf x}}
\def \y{{\bf y}}
\def \z{{\bf z}}
\def \b{{\bf b}}
\def \d{{\bf d}}
\def \c{{\bf c}}
\def \a{{\bf a}}
\def \al{{\bm{\alpha}}}
\def \f{{\bf f}}
\def \i{{\bf i}}
\def \o{{\bf o}}
\def \hw{\hat{\bf{w}}}
\def \W{{\bf W}}
\def \D{{\bf D}}
\def \H{{\bf H}}
\def \X{{\bf X}}
\def \I{{\bf I}}

\def \hW{{\hat{\bf{W}}}}
\def \m{{\bf m}}
\def \v{{\bf v}}
\def \s{{\bf s}}

\def\l{\lambda}
\def\diag{\text{diag}}

\newcommand{\NM}[2 ]{\left\| #1 \right\|_{#2}}

\begin{document}

\maketitle

\begin{abstract}
Deep neural network models, though very powerful and highly successful, are computationally expensive in 
terms of space and time.
Recently, there have been a number of attempts on binarizing the network weights and
activations. This greatly reduces the network size, and replaces the underlying
multiplications to additions or even XNOR bit operations. However, existing
binarization schemes are based on simple matrix approximations and ignore the effect of
binarization 
on the loss. 
In this paper, we 
propose a proximal Newton algorithm with diagonal Hessian approximation that
directly minimizes the loss
w.r.t. the binarized weights. 
The underlying proximal step has an efficient closed-form solution, and
the second-order information can be efficiently obtained from the second moments already
computed by 
the Adam optimizer.
Experiments on both feedforward and recurrent networks
show that the proposed loss-aware binarization algorithm outperforms existing
binarization schemes,
and is also more robust for wide and deep networks. 
\end{abstract}


\section{Introduction}



Recently, deep neural networks have achieved state-of-the-art performance in
various tasks such as speech recognition, visual object recognition, and image
classification~\citep{lecun2015deep}. Though powerful,
the large number of network weights  leads to space and time inefficiencies in 
both training and storage.
For instance, the popular AlexNet, VGG-16 and Resnet-18 all
require hundred of megabytes to store, and billions of high-precision
operations on classification. 
This limits its use in embedded systems, smart phones and other portable devices
that are now everywhere.

To alleviate this problem,
a number of approaches have been recently proposed.
One attempt first trains a neural network and then compresses it
\citep{han2015deep,kim2015compression}.  Instead of this two-step approach, it is more
desirable to train and compress the network simultaneously. Example approaches include
tensorizing 
\citep{novikov2015tensorizing}, parameter quantization
\citep{gong2014compressing}, and 
binarization~\citep{courbariaux2015binaryconnect,hubara2016binarized,rastegari2016xnor}.
In particular,
binarization only requires one bit for each weight value. This can significantly reduce
storage, and also eliminates most multiplications during the forward pass.


\citet{courbariaux2015binaryconnect} pioneered neural network binarization with 
the BinaryConnect algorithm,
which achieves state-of-the-art results on many
classification tasks.
Besides binarizing the weights, \citet{hubara2016binarized} further binarized the activations.
\citet{rastegari2016xnor} 
also learned to scale the binarized weights,
and obtained better results.
Besides, they proposed the XNOR-network with both weights and activations binarized as in 
\citep{hubara2016binarized}.
Instead of binarization,
ternary-connect quantizes each weight to 
$\{-1,0,1\}$
\citep{lin2015neural}.
Similarly, the ternary weight network 
\citep{li2016ternary}
and
DoReFa-net~\citep{zhou2016dorefa} 
quantize weights to 
three levels or more.
However, though using more bits allows more accurate weight
approximations, specialized hardwares are needed for the underlying
non-binary operations.




Besides the huge amount of computation and storage involved, 
deep networks are  
difficult to train because of the
highly nonconvex
objective 
and inhomogeneous curvature.
To alleviate this problem,
Hessian-free methods~\citep{martens2012training} 
use the second-order information 
by conjugate gradient.
A related method is
natural gradient descent
\citep{pascanu2013revisiting}, 
which utilizes geometry of the underlying parameter manifold.
Another approach 
uses element-wise
adaptive learning rate,
as in Adagrad \citep{duchi2011adaptive}, 
Adadelta~\citep{zeiler2012adadelta}, RMSprop~\citep{tieleman2012lecture}, 
and Adam~\cite{kingma2014adam}.
This can also be considered as preconditioning that rescales
the gradient so that all dimensions have similar curvatures. 


In this paper, instead of directly approximating the weights,
we propose to consider the effect of binarization on the loss during
binarization. We formulate this as an optimization problem using the proximal Newton algorithm \citep{lee2014proximal} with a diagonal Hessian. The crux of proximal algorithms is the proximal step.
We show that this step has a closed-form solution, whose form is similar to the use of element-wise adaptive learning rate. 
The proposed method also reduces to 
BinaryConnect 
\citep{courbariaux2015binaryconnect} 
and the Binary-Weight-Network 
\citep{hubara2016binarized} 
when curvature information is dropped.
Experiments on both feedforward and recurrent neural
network models show that it outperforms existing binarization algorithms.
In particular, BinaryConnect fails on deep recurrent networks because of the exploding
gradient problem, while the proposed method still demonstrates robust performance.


\textbf{Notations:}
For a vector $\x$,
$\sqrt{\x}$ denotes the element-wise square root,
$|\x|$ denotes the element-wise absolute value,
$\|\x\|_p = (\sum_i|x_i|^p)^{\frac{1}{p}}$ is the $p$-norm of $\x$, 
$\x \succ 0$ denotes that all entries of $\x$ are positive, 
$\text{sign}(\x)$ is the vector with 
$[\text{sign}(\x)]_i
=1$ if $x_i \geq 0$ and $-1$ otherwise,
and
$\text{Diag}(\x)$ returns a  diagonal matrix with
$\x$
on the diagonal.
For two vectors $\x$ and $\y$,
$\x \odot \y$ denotes the element-wise multiplication
and $\x \oslash \y$ denotes the element-wise division.
For a matrix $\X$,
$\text{vec}(\X)$ returns the vector obtained by stacking 
the columns 
of $\X$,
and $\text{diag}(\X)$ returns a diagonal matrix whose diagonal elements are extracted from diagonal of $\X$.


\section{Related Work}


\subsection{Weight Binarization in Deep Networks}
	

In a feedforward neural network with $L$ layers, let the weight matrix 
(or tensor in the case of a convolutional layer) at layer $l$
be $\W_l$.
We combine 
the (full-precision) weights from all layers as $\w =
[\w_1^\top,  \w_2^\top, \dots, \w_L^\top]^\top$, where
$\w_l =  \text{vec}(\W_l)$.
Analogously,
the binarized weights 
are denoted
as
$\hw=[\hw_1^\top, \hw_2^\top, \dots, \hw_L^\top]^\top$.
As 
it is essential to
use full-precision weights during updates 
\citep{courbariaux2015binaryconnect}, 
typically 
binarized weights are only used 
during the forward and backward propagations, but
not on parameter update.
At the $t$th iteration, 
the (full-precision) weight $\w_l^t$ is updated by using
the backpropagated gradient 
$\nabla_l \ell(\hw^{t-1})$ (where
$\ell$ is 
the loss and $\nabla_l \ell(\hw^{t-1})$ is the partial derivative of $\ell$ w.r.t. the 
weights of the $l$th layer).
In the next forward propagation, 
it is then binarized as
$\hw_l^t = \text{Binarize}(\w_l^t)$, where
$\text{Binarize}(\cdot)$ is some binarization scheme.



The two most popular binarization schemes are 
BinaryConnect \citep{courbariaux2015binaryconnect} and
Binary-Weight-Network (BWN) \citep{rastegari2016xnor}.
In
BinaryConnect,
binarization is performed by 
transforming
each element of $\w_l^t$ 
to $-1$ or $+1$ using the sign function:\footnote{A stochastic binarization scheme is also
proposed in \citep{courbariaux2015binaryconnect}.  However, it is much more computational
expensive than (\ref{eq:bc}) and so will not be considered here.}
\begin{equation}\label{eq:bc}
\text{Binarize}(\w_l^t) = \text{sign}(\w_l^t).
\end{equation}
Besides the binarized weight matrix, a scaling parameter 
is also learned in
BWN.
In other words, $\text{Binarize}(\w_l^t) = \alpha_l^t \b_l^t$,
where $\alpha_l^t>0$ and $\b_l^t$ is binary.  They
are obtained by minimizing the difference between
$\w_l^t$ and $\alpha_l^t \b_l^t$,
and have a simple closed-form solution:
\begin{equation} \label{eq:bwn}
\alpha_l^t = \frac{\|\w_l^t\|_1}{n_l},
\;\; 
\b_l^t = \text{sign} (\w_l^t),
\end{equation}
where $n_l$  is  the number of weights in layer $l$. 
\citet{hubara2016binarized} further binarized the activations
as $\hat{\x}_l^t = \text{sign}(\x_l^t)$,
where $\x_l^t$ is the activation of the $l$th layer at iteration $t$.


\subsection{Proximal Newton Algorithm}
\label{sec:prox}
	
The proximal Newton algorithm \citep{lee2014proximal} has been popularly used for solving
composite optimization problems of the form
\[ \min_{\x} f(\x) + g(\x), \]
where $f$ is 
convex
and 
smooth, and $g$ is convex but possibly nonsmooth.
At iteration $t$, it
generates the next iterate as
\[ \x_{t + 1} = \arg\min_{\x} 
\nabla f(\x_t)^\top 
(\x - \x_t) + (\x- \x_t)^{\top} \H (\x - \x_t) + g(\x), \]
where $\H$ is an approximate Hessian matrix of $f$ at $\x_t$.
With the use of second-order information, the proximal Newton algorithm
converges faster than the proximal gradient algorithm \citep{lee2014proximal}.
Recently, by assuming that $f$ and $g$ have difference-of-convex decompositions
\citep{yuille2002concave},
the proximal Newton algorithm is also extended to the case where $g$ is nonconvex
\citep{rakotomamonjy2016dc}.


\section{Loss-Aware Binarization}

As can be seen, existing weight binarization methods 
\citep{courbariaux2015binaryconnect,rastegari2016xnor}
simply find the closest binary approximation of $\w$, and ignore
its effects to the loss.
In this paper, we consider the loss directly during binarization.
As in \citep{rastegari2016xnor},
we also binarize the weight 
$\w_l$  
in each layer 
as $\hw_l =\alpha_l \b_l$, where $\alpha_l>0$  and $\b_l$ is binary.

In the following, we make the following assumptions on $\ell$.
(A1) $\ell$ is continuously differentiable with Lipschitz-continuous gradient,
	i.e., there exists $\beta>0$ such that 
	$\NM{\nabla \ell(\u) - \nabla \ell(\v)}{2} 
	\leq \beta \NM{\u-\v}{2}$ for any $\u, \v$;
(A2) $\ell$ is bounded from below.


\subsection{Binarization using Proximal Newton Algorithm}

We formulate
weight binarization 
as the following optimization problem:
\begin{eqnarray}
& \min_{\hw} &  \ell(\hw) \label{eq:obj}\\
& \text{s.t.} &  
\hw_l=\alpha_l \b_l, \;
\alpha_l > 0,
\;
\b_l\in \{\pm 1\}^{n_l}, \;\; l=1,\dots,L, \label{eq:constraints}
\end{eqnarray}
where $\ell$ is the loss.
Let $C$ be the feasible region in (\ref{eq:constraints}), and define
its indicator function:
$I_C(\hw) =  
0$ if $\hw\in C$, and
$\infty$ otherwise.
Problem~(\ref{eq:obj}) can then be rewritten as
\begin{equation} \label{eq:obj2}
\min_{\hw}
\; \ell(\hw) + I_C(\hw). 
\end{equation}
We solve
(\ref{eq:obj2}) using the proximal Newton method (Section~\ref{sec:prox}).
At iteration $t$,
the smooth term
$\ell(\hw^t)$ is replaced by the second-order expansion
\[ 
\ell(\hw^{t-1}) +  \nabla \ell(\hw^{t-1})^\top (\hw^t -
\hw^{t-1}) + \frac{1}{2}(\hw^t - \hw^{t-1})^\top \H^{t-1} (\hw^t - \hw^{t-1}),
\]
where $\H^{t-1}$ is an estimate of the Hessian of $\ell$ at $\hw^{t-1}$.
Note that using the Hessian to capture second-order information is essential for efficient
neural network training, as
$\ell$ is often flat in some directions but highly curved in others.
By rescaling the gradient, the loss has similar curvatures along all directions.
This is also called preconditioning in the literature \citep{dauphin2015equilibrated}. 

For neural networks, the exact Hessian is rarely positive semi-definite. This can be problematic as the
nonconvex objective leads to indefinite quadratic optimization.
Moreover, 
computing the exact Hessian is both time- and space-inefficient on large networks. To alleviate
these problems, a popular
approach is to 
approximate the Hessian by a diagonal positive definite matrix $\D$.
One popular choice is the efficient Jacobi preconditioner.
Though an efficient approximation of the
Hessian under certain conditions, it is not
competitive for indefinite matrices \citep{dauphin2015equilibrated}.
More recently, 
it is shown that equilibration provides a more robust preconditioner 
in the presence of saddle points
\citep{dauphin2015equilibrated}. This is also adopted by popular stochastic optimization
algorithms such as RMSprop
\citep{tieleman2012lecture} 
and Adam
\citep{kingma2014adam}.
Specifically,
the second moment $\v$ in these algorithms is an estimator of
$\diag(\H^2)$
\citep{dauphin1502rmsprop}.
Here, we 
use the square root of this $\v$, which is readily available in Adam, to construct 
	$\D = \text{Diag}([\diag(\D_1)^\top, \dots, \diag(\D_L)^\top]^\top)$,
where $\D_l$ is the approximate diagonal Hessian 
at layer
$l$.
In general, other estimators of 
$\diag(\H)$
 can also be used.

At the $t$th 
iteration
of the proximal Newton algorithm, the  following subproblem
is solved:
\begin{eqnarray} \label{eq: obj_proximal}
&\min_{\hw^t}
&   \nabla \ell(\hw^{t-1})^\top (\hw^t - \hw^{t-1}) + \frac{1}{2}(\hw^t -
\hw^{t-1})^\top \D^{t-1} (\hw^t - \hw^{t-1})  \\ \nonumber
& \text{s.t.} &  \hw_l^t=\alpha_l^t \b_l^t, 
\; \alpha_l^t > 0, \; \b_l^t\in \{\pm 1\}^{n_l},
\quad l = 1, \dots, L.
\nonumber
\end{eqnarray}

\begin{prop} \label{prop:opt}
Let
$\d^{t-1}_l \equiv \text{diag}(\D^{t-1}_l)$, and 
\begin{equation} \label{eq:wt}
\w^t_l \equiv \hw_l^{t-1} - \nabla_l \ell(\hw^{t-1}) \oslash \d^{t-1}_l.
\end{equation} 
The optimal solution of (\ref{eq: obj_proximal}) can be obtained in closed-form as 
\begin{equation} \label{eq:opt}
\alpha^t_l = \frac{\|\d^{t-1}_l \odot \w^t_l\|_1}{\|\d^{t-1}_l\|_1}, \;\;
\b_l^t = \text{sign}(\w^t_l).
\end{equation}
\end{prop}

\begin{theorem} \label{thm:conv}
Assume that 
$[\d_l^t]_k > \beta \;
\forall l,k,t$, the objective of \eqref{eq:obj2} produced by the proximal Newton algorithm (with
closed-form update of $\hw^t$ in Proposition~\ref{prop:opt}) converges. 
\end{theorem}

Note that both the loss $\ell$ and indicator function $I_C(\cdot)$ in \eqref{eq:obj2} are
not convex.  Hence, convergence analysis of the proximal Newton algorithm in \citep{lee2014proximal}, which is only for convex problems, cannot be applied.
Recently, \cite{rakotomamonjy2016dc} proposed a nonconvex proximal Newton extension. However, 
it assumes a difference-of-convex decomposition 
which does not
hold  here.

\begin{remark}
When
$\D_l^{t-1} = \lambda \I$,
i.e., the curvature is the same for all
	dimensions in the $l$th layer, 
(\ref{eq:opt}) then reduces to 
the BWN solution in (\ref{eq:bwn})
In other words, BWN corresponds to using the proximal
gradient algorithm, while the proposed method corresponds to the proximal Newton
algorithm with diagonal Hessian. 
In composite optimization, it is known that the proximal Newton method is more efficient than the
proximal gradient algorithm~\citep{lee2014proximal,rakotomamonjy2016dc}. 

\end{remark}

\begin{remark}
	When $\alpha^t_l=1$,
	(\ref{eq:opt}) reduces to 
	$\text{sign}(\w_l^t)$, which is the 
	BinaryConnect solution in (\ref{eq:bc}).
\end{remark}

From 
(\ref{eq:wt}) and (\ref{eq:opt}),
each iteration first performs gradient descent along $ \nabla_l \ell(\hw^{t-1})$ with an
adaptive learning rate $1 \oslash \d^{t-1}_l$, and then projects it to a binary solution.
As discussed in \citep{courbariaux2015binaryconnect},
it is important to 
keep a full-precision weight during training.  Hence, we replace (\ref{eq:wt}) by
$\w^t_l  \leftarrow \w^{t-1}_l  -  \nabla_l \ell(\hw^{t-1}) \oslash \d_l^{t-1} $.
The whole procedure, which will be called
Loss-Aware Binarization (LAB),
is shown in Algorithm~\ref{alg:3}. 
In steps 5 and 6, following~\citep{li2016ternary},
we first rescale input 
$\x_l^{t-1}$ 
to the $l$th layer  
with $\alpha_l$, so that multiplications in dot products
and convolutions 
become additions. 



\begin{algorithm}[h]
	\caption{Loss-Aware Binarization (LAB)
	for training a feedforward neural network.}\label{alg:3}
	\textbf{Input: }Minibatch 
	$\{(\x_0^t,\y^t)\}$,  current full-precision weights $\{\w^t_l\}$,
	first moment $\{\m^{t-1}_l\}$,
	second moment $\{\v^{t-1}_l\}$,
	and learning rate $\eta^t$. 
	\begin{algorithmic}[1]
		\STATE {\bf Forward Propagation}
		\FOR{$l=1$ to $L$}
		\STATE $\alpha^t_l =  \frac{\|\d^{t-1}_l \odot \w^t_l\|_1}{\|\d^{t-1}_l\|_1}$;
		\STATE $\b^t_l = \text{sign}(\w^t_l)$;
		\STATE rescale the layer-$l$ input: $\tilde{\x}^t_{l-1} = \alpha^t_l \x^t_{l-1}$;
		\STATE compute $\z^t_l$ with input  $\tilde{\x}^t_{l-1}$ and binary weight $\b^t_l$; 
		\STATE apply batch-normalization 
		and nonlinear activation to $\z^t_l$ to obtain $\x^t_l$; 
		\ENDFOR
		\STATE compute the loss $\ell$ using $\x^t_{L}$ and $\y^t$;
		\STATE {\bf Backward Propagation}
		\STATE initialize output layer's activation's gradient $\frac{\partial \ell}{\partial \x^t_{L}}$;
		\FOR{$l=L$ to $2$}
		\STATE compute $\frac{\partial \ell}{\partial \x^t_{l-1}}$ using $\frac{\partial
			\ell}{\partial \x^t_l}$, $\alpha^t_l$ and $\b^t_l$;
		\ENDFOR
		\STATE {\bf Update parameters using Adam}
		\FOR{$l=1$ to $L$}
		\STATE compute gradients $\nabla_l \ell(\hw^t)$ using $\frac{\partial \ell}{\partial \x^t_l}$ and $\x^t_{l-1}$;
		\STATE update first moment $\m^t_l = \beta_1 \m^{t-1}_l + (1-\beta_1)\nabla_l
		\ell(\hw^t)$;
		\STATE update second moment $\v^t_l = \beta_2 \v^{t-1}_l + (1-\beta_2)(\nabla_l \ell(\hw^t) \odot \nabla_l \ell(\hw^t))$;
		\STATE compute unbiased first moment $\hat{\m}^t_l = \m^t_l/(1-\beta_1^t)$;
		\STATE compute unbiased second moment $\hat{\v}^t_l = \v^t_l/(1-\beta_2^t)$;
		\STATE compute current curvature matrix $\d^t_l = \frac{1}{\eta^t}\left(\epsilon \bm{1}+\sqrt{\hat{\v}^t_l}
		\right)$;
		\STATE update full-precision weights $\w^{t+1}_l = \w^t_l -  \hat{\m}^t_l \oslash \d^t_l$;
		\STATE update learning rate $\eta^{t+1} = \text{UpdateRule} (\eta^t, t+1)$;
		\ENDFOR
	\end{algorithmic}
\end{algorithm}

While binarizing weights changes most multiplications to additions,
binarizing both weights and activations 
saves even more computations as
additions 
are further changed
to XNOR bit operations
\citep{hubara2016binarized}.
Our Algorithm~\ref{alg:3} can also be easily extended 
by binarizing the activations with the simple sign function.






\subsection{Extension to Recurrent Neural Networks}
\label{sec:rnn_theory}


The proposed method can be easily extended to recurrent neural networks.
Let $\x_l$ and 
$\h_l$ 
be the input and 
hidden states, respectively, at time step (or depth)
$l$.
A typical recurrent neural network
has a recurrence of the form 
$\h_l = \W_{x}\x_l  + \W_{h} \sigma(\h_{l-1}) + \b$ (equivalent to the more widely known  $\h_l = \sigma(\W_{x}\x_l  + \W_{h}\h_{l-1}   + \b)$~\citep{pascanu2013difficulty} ).
We binarize both 
the input-to-hidden weight $\W_x$
and 
hidden-to-hidden weight $\W_h$.
Since weights are shared across time in a recurrent network, we only need to binarize
$\W_{x}$ and $\W_{h}$ once in each forward propagation.
Besides weights,
one can also binarize the activations (of the 
inputs and
hidden states) as in the previous section.


In deep networks, 
the backpropagated gradient takes the form of a product of Jacobian
matrices~\citep{pascanu2013difficulty}.
In a vanilla recurrent neural network,\footnote{Here, we consider the vanilla
recurrent neural network for simplicity. It can be shown that a similar behavior holds for
the more commonly used LSTM.}
for activations 
$\h_p$ and $\h_q$
at depths $p$ and $q$, respectively (where $p>q$), 
$\frac{\partial{\h_p}}{\partial\h_q} = \prod_{q<l\leq p}
\frac{\partial{\h_l}}{\partial\h_{l-1}} = \prod_{q<l\leq p} \W_h^\top
\text{diag}(\sigma'(\h_{l-1}))$.
The
necessary condition for exploding
gradients is that the largest singular value 
$\lambda_1(\W_h)$
of 
$\W_h$
is larger than some given constant
\citep{pascanu2013difficulty}.
The following Proposition shows that
for any binary $\W_h$,
its largest singular value 
is lower-bounded by
the square root of its dimension.

\begin{prop}\label{pr:singular}
	For any
	$\W \in \{-1, +1\}^{m\times n}$ ($m \le n$), 
	$\lambda_1(\W) \geq \sqrt{n}$. 
\end{prop}
Thus, with weight binarization as in BinaryConnect, the exploding gradient problem 
becomes more severe as
the weight matrices are often large.
On the other hand,
recall that $\lambda_1(c \hW_h) = c \lambda_1(\hW_h)$ for any non-negative $c$.
The proposed method alleviates this 
exploding
gradient problem by 
adaptively learning
the scaling parameter $\alpha_h$.

\section{Experiments}
\label{sec:expt}

In this section, we perform experiments on the proposed binarization scheme with both
feedforward networks 
(Sections~\ref{sec:fnn} and \ref{sec:fnn1}) and 
recurrent neural networks
(Sections~\ref{sec:rnn} and \ref{sec:rnn1}).


\subsection{Feedforward Neural Networks}
\label{sec:fnn}

We compare the original full-precision network (without binarization)
with the
following weight-binarized networks:
(i) BinaryConnect;
(ii) Binary-Weight-Network
(BWN);
and
(iii) the proposed Loss-Aware Binarized network (LAB).
We also compare with 
networks having both
weights and activations binarized:\footnote{We use the straight-through-estimator \citep{hubara2016binarized} to compute
the gradient involving the sign function.}
(i) BinaryNeuralNetwork (BNN)~\citep{hubara2016binarized}, the weight-and-activation binarized counterpart of BinaryConnect;
(ii) XNOR-Network (XNOR)~\citep{rastegari2016xnor}, the 
counterpart of BWN;
(iii) LAB2,
the 
counterpart of the
proposed method, which
binarizes weights using proximal Newton method and binarizes activations using a simple sign function.

The setup is similar to that in \citet{courbariaux2015binaryconnect}.
We do not perform data augmentation
or unsupervised pretraining.
Experiments are performed on three commonly used data sets:
\begin{enumerate}
\item  \emph{MNIST}:
This contains $28 \times 28$ gray images from ten digit classes.  We use $50000$ images for training,
another $10000$ for validation, and the remaining $10000$  for testing.  We use the 4-layer model:
\[ 784FC-2048FC-2048FC-2048FC-10SVM, \]
where $FC$ is a fully-connected layer, and $SVM$ is a L2-SVM output layer using the square hinge
	loss.
Batch normalization, with a minibatch size $100$,
is used to accelerate learning. 
The maximum number of epochs is $50$.
The learning rate for the weight-binarized (resp. weight-and-activation-binarized) network
starts at $0.01$ (resp. $0.005$),
 and decays by a factor of $0.1$ at epochs $15$ and $25$.
	\item \emph{CIFAR-10}:
	This contains 
	$32 \times 32$ color images from
	ten object classes.
	We 
	use $45000$ images for training, another
	$5000$
	for validation,
	and the remaining
	$10000$ 
	for testing. 
The images are preprocessed with global contrast normalization and ZCA whitening.
		We 
	use the VGG-like architecture:
	\[
	(2\times 128C3)-MP2-(2 \times 256C3)-MP2-(2 \times 512C3)-MP2-(2 \times 1024FC)-10SVM,
	\]
	where $C3$ is a $3 \times 3$ ReLU convolution layer, and $MP2$ is a $2 \times 2$ max-pooling layer. 
	Batch normalization, with a minibatch size of $50$, is used.
	The maximum number of epochs is $200$.
The learning rate for the weight-binarized (resp. weight-and-activation-binarized) network
starts at $0.03$ (resp. $0.02$), and decays by a factor of $0.5$ after every 15 epochs. 
	
	\item \emph{SVHN}:
	This contains
	$32\times 32$ color images
	from ten digit classes.
	We use $598388$ images for training, another
	$6000$ for validation,
	and the remaining
	$26032$ 
	for testing.
The images are preprocessed with global and local contrast normalization.
	The model used is:
	\[
	(2\times 64C3)-MP2-(2 \times 128C3)-MP2-(2 \times 256C3)-MP2-(2 \times 1024FC)-10SVM. 
	\]
	Batch normalization, with a minibatch size of $50$, is used.
	The maximum number of epochs is $50$. 
The learning rate for the weight-binarized (resp. weight-and-activation-binarized) network starts at
$0.001$ (resp. $0.0005$),
	and decays by a factor of $0.1$ at epochs $15$ and $25$.
\end{enumerate}

Since binarization is a form of regularization
\citep{courbariaux2015binaryconnect}, we do not use other regularization methods (like Dropout).
All the weights are 
initialized as in \citep{glorot2010understanding}. 
Adam \citep{kingma2014adam} is used as the optimization solver.

\begin{figure}[htbp]
	\vspace{-10px}
	\begin{center}
		\subfigure[MNIST. \label{fig:MNIST}]{\includegraphics[width=0.29\textwidth]{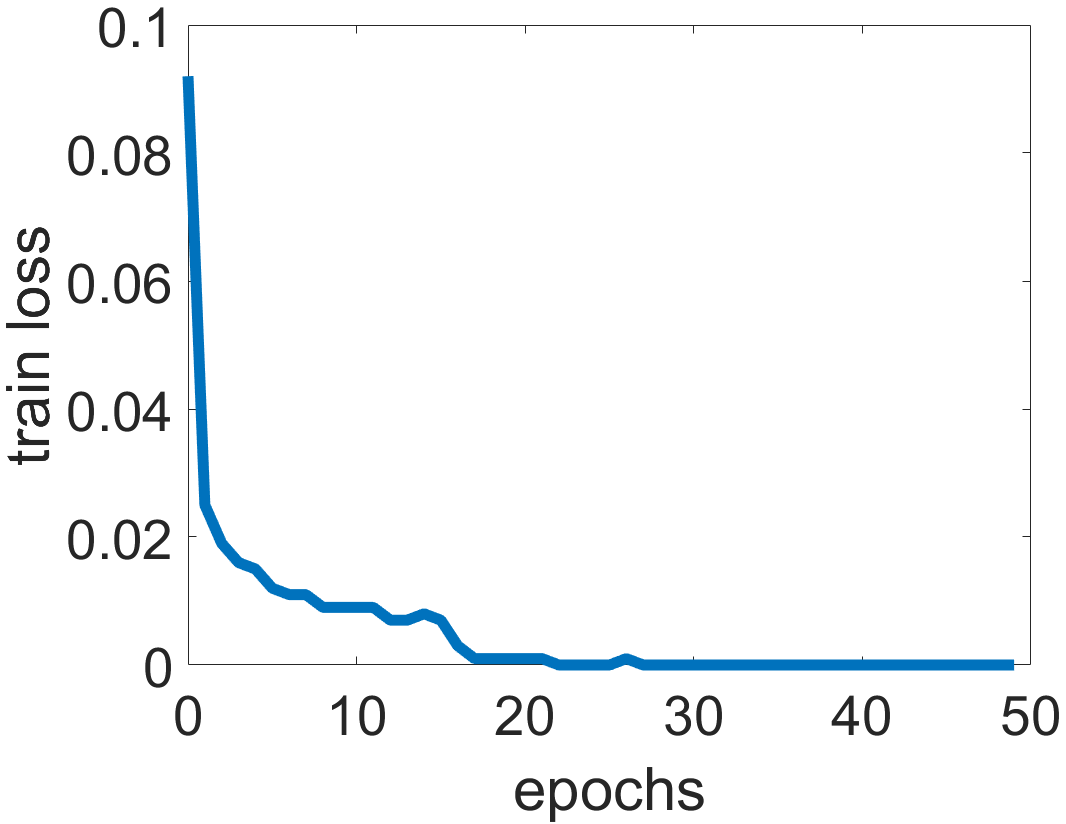}}
		\subfigure[CIFAR-10.\label{fig:CIFAR}]{\includegraphics[width=0.29\textwidth]{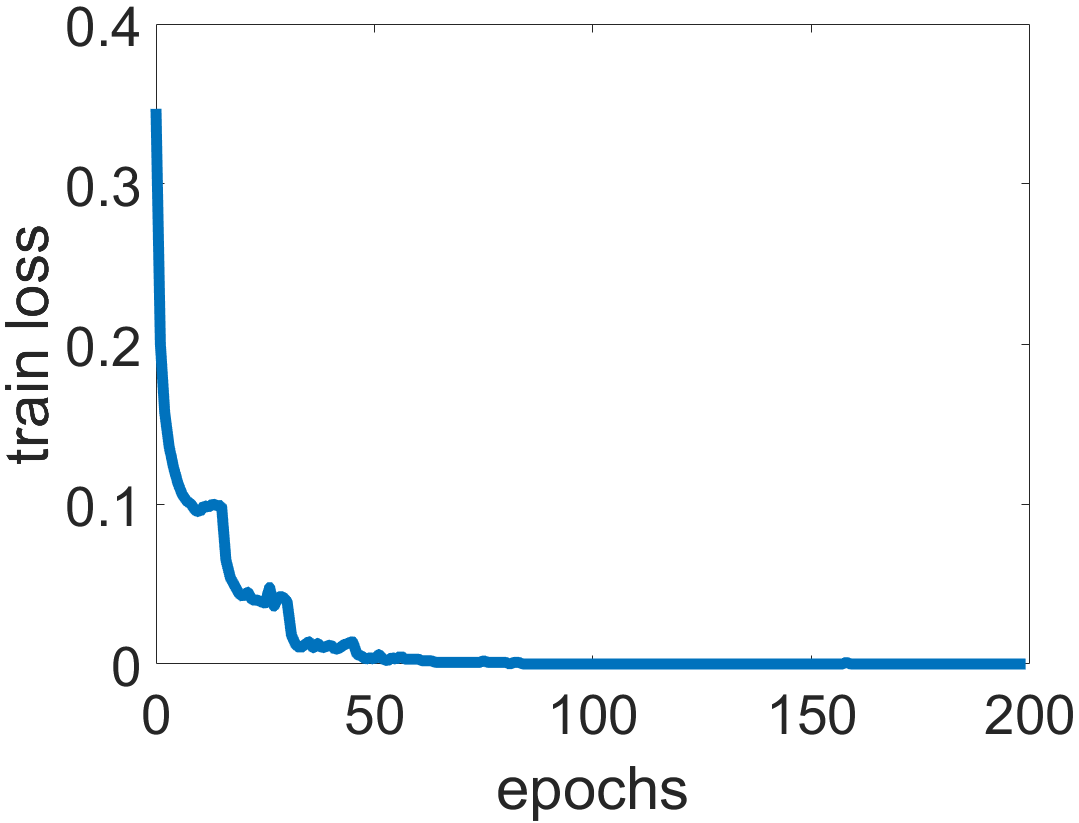}}
		\subfigure[SVHN.\label{fig:SVHN}]{\includegraphics[width=0.29\textwidth]{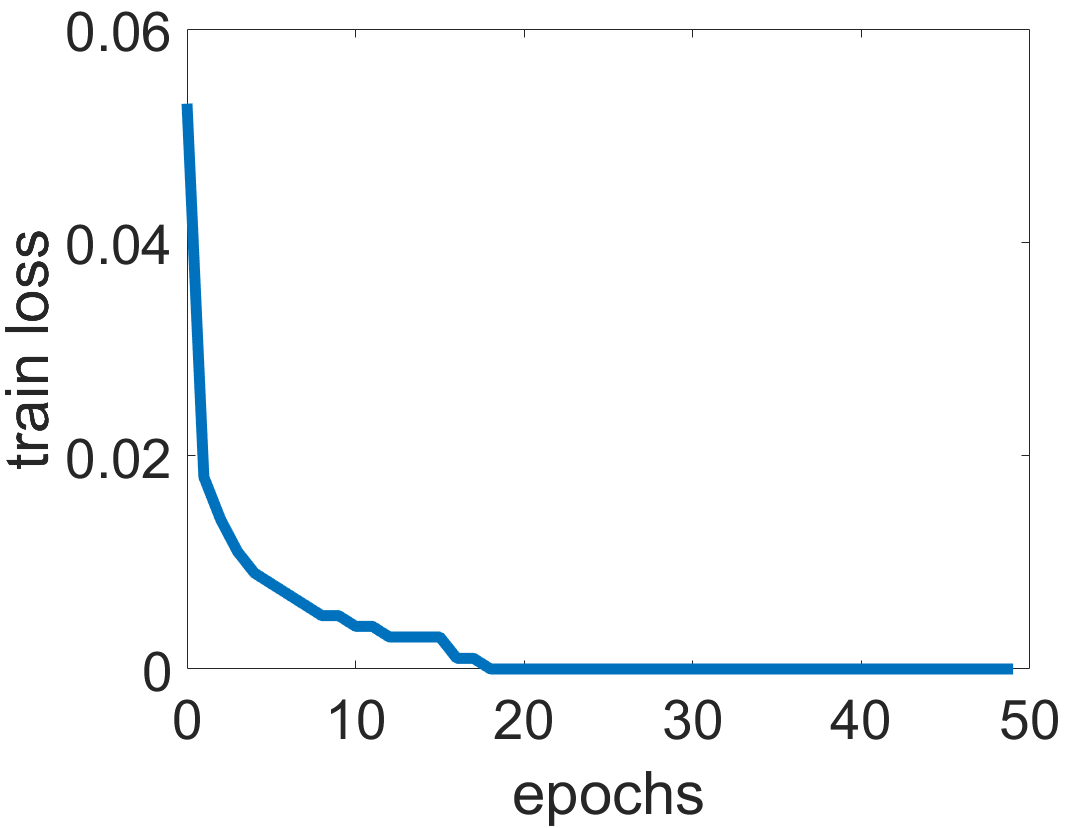}}
\vspace{-.1in} 
\caption{Convergence of LAB with feedforward neural networks.}
\label{fig:training curve}
\end{center}
\end{figure}

Table~\ref{tbl:fnn}
shows
the test classification error rates, and
Figure~\ref{fig:training curve} shows the convergence of LAB.
As can be seen,
the proposed LAB achieves the lowest error
on \emph{MNIST} and \emph{SVHN}.
It even outperforms the full-precision network on \emph{MNIST}, as 
weight binarization serves as a regularizer.
With the use of curvature information,
LAB outperforms BinaryConnect and BWN.
On \emph{CIFAR-10}, 
LAB is slightly outperformed by 
BinaryConnect,
but
is still better than the full-precision network. 
Among the schemes that
binarize
both weights and activations,
LAB2 also outperforms BNN and the XNOR-Network.

\begin{table}[htbp]
\centering
\vspace{-10px}
\caption{Test error rates (\%) for feedforward neural network models.}
\label{tbl:fnn}
\begin{tabular}{cc|c|c|c}
	\hline
	                                   &                &  \emph{MNIST}  & \emph{CIFAR-10} &  \emph{SVHN}   \\ \hline
	        (no binarization)          & full-precision &     1.190      &     11.900      &     2.277      \\ \hline\hline
	                                   & BinaryConnect  &     1.280      & \textbf{9.860}  &     2.450      \\ \cline{2-5}
	        (binarize weights)         &      BWN       &     1.310      &     10.510      &     2.535      \\ \cline{2-5}
	                                   &      LAB       & \textbf{1.180} &     10.500      & \textbf{2.354} \\ \hline\hline
	                                   &      BNN       &     1.470      &     12.870      &     3.500      \\ \cline{2-5}
	(binarize weights and activations) &      XNOR      &     1.530      &     12.620      &     3.435      \\ \cline{2-5}
	                                   &      LAB2      & \textbf{1.380} & \textbf{12.280} & \textbf{3.362} \\ \hline
\end{tabular}
\end{table}


\subsection{Varying the Number of Filters in CNN}
\label{sec:fnn1}

As in \citet{zhou2016dorefa},
we study sensitivity to network width by varying the number of filters 
{\sf K} 
on the \emph{SVHN} data set.
As in Section~\ref{sec:fnn},
we use the model 
\[
(2\times {\sf K} C3)-MP2-(2 \times 2{\sf K} C3)-MP2-(2 \times 4{\sf K} C3)-MP2-(2 \times
1024FC)-10SVM. \]
Results are shown in Table~\ref{tbl:width}.
Again, the proposed LAB has the best performance.
Moreover,
as the number of filters
increases, degradation due to binarization
becomes less severe. 
This suggests that more powerful models 
(e.g., CNN with more filters, standard feedforward networks with 
more hidden units)
are less susceptible to performance degradation due to binarization. 
We speculate that this is because large networks often have larger-than-needed capacities,
and so are less affected by the limited expressiveness of binary weights. Another related reason is
that binarization acts as regularization, and so contributes positively to the
performance.


\begin{table}[htbp]
	\centering
	\vspace{-10px}
\caption{Test error rates (\%) on \emph{SVHN}, for CNNs with different numbers of
filters. Number in brackets is the difference between the errors of the binarized scheme and the full-precision network.}
	\label{tbl:width}
	\begin{tabular}{ c|c|c|c|c}
		\hline
		               &     ${\sf K} = 16$     &     ${\sf K} = 32$     &     ${\sf K} = 64$     &    ${\sf K} = 128$     \\ \hline
		full-precision &         2.738          &         2.585          &         2.277          &         2.146          \\ \hline\hline
		BinaryConnect  &     3.200  (0.462)     &  2.777       (0.192)   &   2.450      (0.173)   &   2.315      (0.169)   \\ \hline
		BWN            &   3.119     (0.461)    &   2.743     (0.158)    &   2.535     (0.258)    &   2.319     (0.173)    \\ \hline
		LAB            & \textbf{3.050} (0.312) & \textbf{2.742} (0.157) & \textbf{2.354} (0.077) & \textbf{2.200} (0.054) \\ \hline
	\end{tabular}
\end{table}


\subsection{Recurrent Neural Networks} 
\label{sec:rnn}

In this section, we perform experiments on the popular
long short-term memory (LSTM) \citep{hochreiter1997long}.
Performance 
is evaluated
in the context of character-level language modeling.
The LSTM takes as input a sequence of characters, and predicts the next character 
at each time step. 
The training objective is the cross-entropy loss 
over all target
sequences.
Following \citet{karpathy2015visualizing},
we use two data sets (with
the same training/validation/test set splitting): 
(i) Leo Tolstoy's \emph{War and Peace},
	which consists of $3258246$ characters of almost entirely English text with minimal markup and has a vocabulary size of $87$; and 
(ii) the source code of the \emph{Linux Kernel}, which
consists of $6206996$ characters and has a vocabulary size of $101$.

We use a one-layer LSTM with $512$ cells.
The maximum number of epochs is $200$, 
and the number of time steps
is $100$.
The initial learning rate 
is 
$0.002$.
After $10$ epochs,
it is decayed 
by a factor of $0.98$ after each epoch.
The weights are initialized uniformly in $[−0.08, 0.08]$.
After each iteration, the gradients are clipped to the range $[-5,5]$, 
and all the updated weights are clipped to $[-1, 1]$. 
For the weight-and-activation-binarized networks, we do not binarize the inputs, as they are one-hot
vectors in this language modeling task.

Table~\ref{tbl:rnn} shows the testing cross-entropy values.
As in Section~\ref{sec:fnn}, the proposed LAB outperforms other weight binarization schemes, and is even
better than the full-precision network on the \emph{Linux Kernel} data set.
BinaryConnect does not work well here
because of the problem of exploding  gradients
(see Section~\ref{sec:rnn_theory} and more results in Section~\ref{sec:rnn1}).
On the other hand, BWN and the proposed LAB
scale the binary weight matrix and
perform better.
LAB 
also performs better than BWN
as
curvature information
is considered.
Similarly, among schemes that
binarize
both weights and activations,
the proposed 
LAB2 also outperforms BNN and XNOR-Network.

\begin{table}[htbp]
\vspace{-10px}
\centering
\caption{Testing cross-entropy values of LSTM.}
\label{tbl:rnn}
\begin{tabular}{ cc|c|c}
	\hline
	                                   &                &  \emph{War and Peace}   & \emph{Linux Kernel} \\ \hline
	(no binarization)                  & full-precision &          1.268          &        1.329        \\ \hline\hline
	                                   & BinaryConnect  &          2.942          &        3.532        \\ \cline{2-4}
	(binarize weights)                 &      BWN       &          1.313          &        1.307        \\ \cline{2-4}
	                                   &      LAB       &     \textbf{1.291}      &   \textbf{1.305}    \\ \hline\hline
	                                   &      BNN       &          3.050          &        3.624        \\ \cline{2-4}
	(binarize weights and activations) &      XNOR      &          1.424          &        1.426        \\ \cline{2-4}
	                                   &      LAB2      &     \textbf{1.376}      &   \textbf{1.409}    \\ \hline
\end{tabular}
\end{table}


\subsection{Varying the Number of Time Steps in LSTM}
\label{sec:rnn1}
\begin{figure}[htbp]
\begin{center}
\subfigure[$TS=10$. \label{fig:depth10}]{\includegraphics[width=0.36\textwidth]{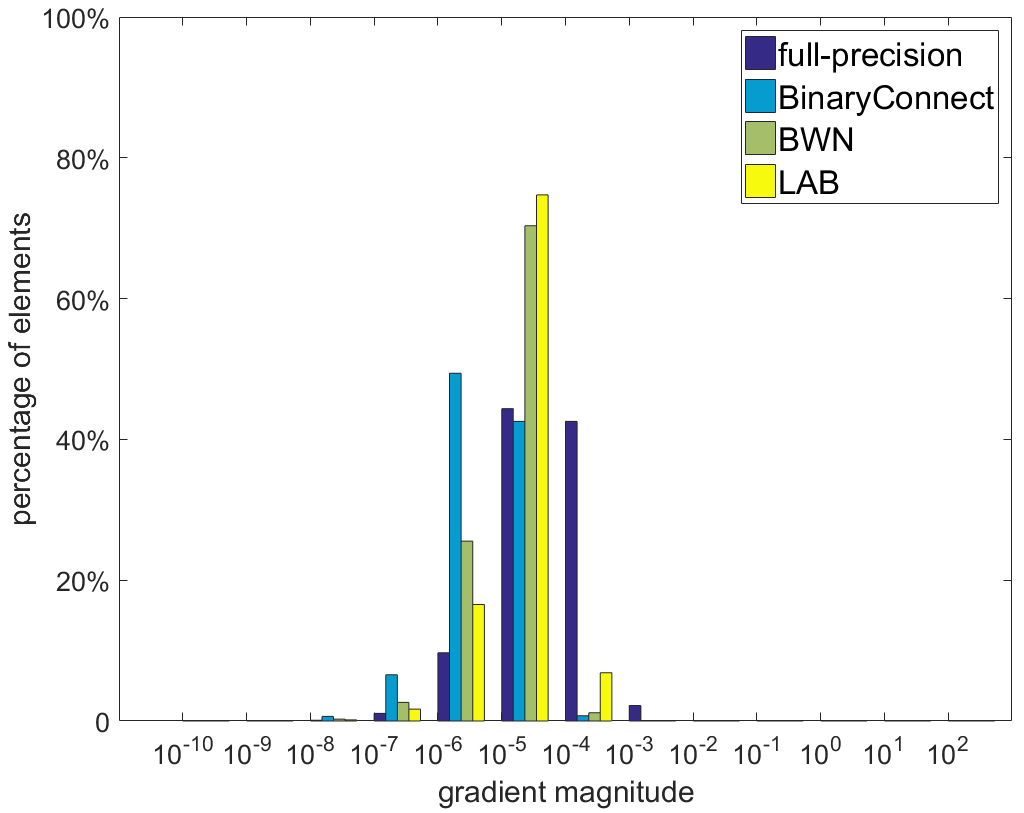}}
\subfigure[$TS=100$.\label{fig:depth100}]{\includegraphics[width=0.36\textwidth]{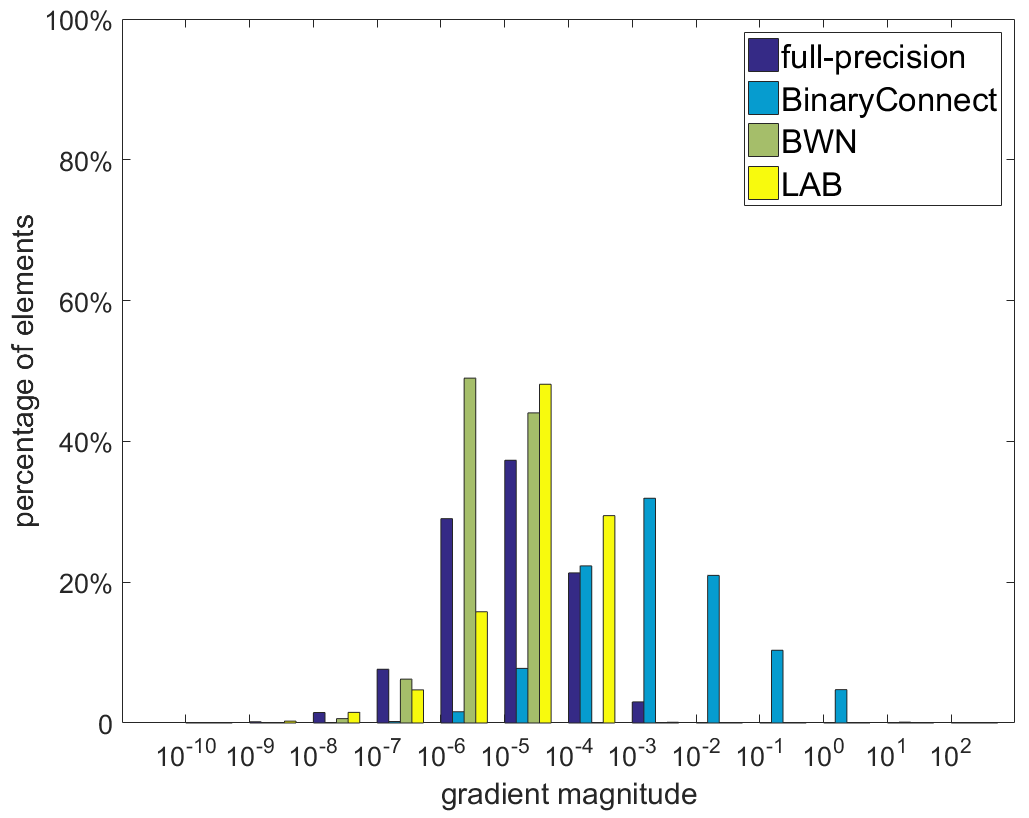}}
\vspace{-.1in} 
\caption{ Distribution of weight gradients on \emph{War and Peace}, for LSTMs with different
time steps.}
		\label{fig:depth}
	\end{center}
\end{figure}

In this experiment,
we study the sensitivity of the binarization schemes with
varying
numbers of unrolled time steps ($TS$)
in LSTM. 
Results are shown in 
Table~\ref{tbl:length}.
Again, the proposed LAB has the best performance. 
When  $TS=10$, the LSTM is relatively shallow,  and all binarization schemes have similar performance
as the full-precision network.
When $TS \geq 50$, 
BinaryConnect fails, while BWN and  the proposed LAB perform better (as discussed in Section~\ref{sec:rnn_theory}).
Figure~\ref{fig:depth} shows
the distributions of the 
hidden-to-hidden weight 
gradients for $TS=10$ and $100$.
As can be seen,  while all models have similar
gradient distributions at $TS=10$, 
the gradient values in BinaryConnect are much higher than those of the other algorithms
for the deeper network ($TS=100$).


\begin{table}[htbp]
	\centering
		\vspace{-10px}
\caption{Testing cross-entropy on \emph{War and Peace}, for LSTMs with different
time steps ($TS$).  
Difference between 
cross-entropies of binarized scheme and full-precision network
is shown in brackets.}
	\label{tbl:length}
	\begin{tabular}{c|c|c|c|c}
		\hline
		               &        $TS=10$         &        $TS=50$         &        $TS=100$        &        $TS=150$        \\ \hline
		full-precision &         1.527          &         1.310          &         1.268          &         1.249          \\ \hline\hline
		BinaryConnect  &   1.528      (0.001)   &   2.980      (1.670)   &   2.942      (1.674)   &  2.872       (1.623)   \\ \hline
		     BWN       &   1.532      (0.005)   &     1.325 (0.015)      &   1.313      (0.045)   &   1.311      (0.062)   \\ \hline
		     LAB       & \textbf{1.527} (0.000) & \textbf{1.324} (0.014) & \textbf{1.291} (0.023) & \textbf{1.285} (0.036) \\ \hline
	\end{tabular}
\end{table}

Note from Table~\ref{tbl:length} that
as the time step increases, all except BinaryConnect
show better performance. However, 
degradation due to binarization also becomes more severe. This is because 
the weights 
are shared across time steps.
Hence, 
error 
due to binarization 
also propagates across time.


\section{Conclusion }

In this paper, we 
propose a 
binarization algorithm that directly considers its effect on the loss during binarization.
The binarized weights are obtained 
using proximal Newton algorithm with diagonal Hessian approximation.
The proximal step has an efficient closed-form solution, and
the second-order information in the Hessian can be readily
obtained from 
the Adam optimizer.
Experiments 
show that the proposed algorithm outperforms existing binarization schemes, has comparable performance
as the original full-precision network, and is also robust for wide and deep networks. 



\subsubsection*{Acknowledgments}

This research was supported in part by the Research Grants Council of the Hong Kong Special
Administrative Region
(Grant 614513).
We thank Yongqi Zhang for helping with the experiments, and
developers of Theano~\citep{2016arXiv160502688short},
Pylearn2~\citep{goodfellow2013pylearn2} and Lasagne.
We also thank NVIDIA for the support of Titan X GPU.

\bibliography{iclr2017_conference}

\begin{thebibliography}{28}
\providecommand{\natexlab}[1]{#1}
\providecommand{\url}[1]{\texttt{#1}}
\expandafter\ifx\csname urlstyle\endcsname\relax
  \providecommand{\doi}[1]{doi: #1}\else
  \providecommand{\doi}{doi: \begingroup \urlstyle{rm}\Url}\fi

\bibitem[Courbariaux et~al.(2015)Courbariaux, Bengio, and
  David]{courbariaux2015binaryconnect}
M.~Courbariaux, Y.~Bengio, and J.P. David.
\newblock {BinaryConnect}: Training deep neural networks with binary weights
  during propagations.
\newblock In \emph{NIPS}, pp.\  3105--3113, 2015.

\bibitem[Dauphin et~al.(2015{\natexlab{a}})Dauphin, de~Vries, and
  Bengio]{dauphin2015equilibrated}
Y.~Dauphin, H.~de~Vries, and Y.~Bengio.
\newblock Equilibrated adaptive learning rates for non-convex optimization.
\newblock In \emph{NIPS}, pp.\  1504--1512, 2015{\natexlab{a}}.

\bibitem[Dauphin et~al.(2015{\natexlab{b}})Dauphin, de~Vries, Chung, and
  Bengio]{dauphin1502rmsprop}
Y.~Dauphin, H.~de~Vries, J.~Chung, and Y.~Bengio.
\newblock {RMSprop} and equilibrated adaptive learning rates for non-convex
  optimization.
\newblock Technical Report arXiv:1502.04390, 2015{\natexlab{b}}.

\bibitem[Duchi et~al.(2011)Duchi, Hazan, and Singer]{duchi2011adaptive}
J.~Duchi, E.~Hazan, and Y.~Singer.
\newblock Adaptive subgradient methods for online learning and stochastic
  optimization.
\newblock \emph{Journal of Machine Learning Research}, 12:\penalty0 2121--2159,
  2011.

\bibitem[Glorot \& Bengio(2010)Glorot and Bengio]{glorot2010understanding}
X.~Glorot and Y.~Bengio.
\newblock Understanding the difficulty of training deep feedforward neural
  networks.
\newblock In \emph{AISTAT}, pp.\  249--256, 2010.

\bibitem[Gong et~al.(2014)Gong, Liu, Yang, and Bourdev]{gong2014compressing}
Y.~Gong, L.~Liu, M.~Yang, and L.~Bourdev.
\newblock Compressing deep convolutional networks using vector quantization.
\newblock Technical Report arXiv:1412.6115, 2014.

\bibitem[Goodfellow et~al.(2013)Goodfellow, Warde-Farley, Lamblin, Dumoulin,
  Mirza, Pascanu, Bergstra, Bastien, and Bengio]{goodfellow2013pylearn2}
I.J. Goodfellow, D.~Warde-Farley, P.~Lamblin, V.~Dumoulin, M.~Mirza,
  R.~Pascanu, J.~Bergstra, F.~Bastien, and Y.~Bengio.
\newblock Pylearn2: a machine learning research library.
\newblock \emph{arXiv preprint arXiv:1308.4214}, 2013.

\bibitem[Han et~al.(2016)Han, Mao, and Dally]{han2015deep}
S.~Han, H.~Mao, and W.J. Dally.
\newblock Deep compression: Compressing deep neural network with pruning,
  trained quantization and {H}uffman coding.
\newblock In \emph{ICLR}, 2016.

\bibitem[Hochreiter \& Schmidhuber(1997)Hochreiter and
  Schmidhuber]{hochreiter1997long}
S.~Hochreiter and J.~Schmidhuber.
\newblock Long short-term memory.
\newblock \emph{Neural Computation}, pp.\  1735--1780, 1997.

\bibitem[Hubara et~al.(2016)Hubara, Courbariaux, Soudry, El-Yaniv, and
  Bengio]{hubara2016binarized}
I.~Hubara, M.~Courbariaux, D.~Soudry, R.~El-Yaniv, and Y.~Bengio.
\newblock Binarized neural networks.
\newblock In \emph{NIPS}, pp.\  4107--4115, 2016.

\bibitem[Karpathy et~al.(2016)Karpathy, Johnson, and
  Li]{karpathy2015visualizing}
A.~Karpathy, J.~Johnson, and F.-F. Li.
\newblock Visualizing and understanding recurrent networks.
\newblock In \emph{ICLR}, 2016.

\bibitem[Kim et~al.(2016)Kim, Park, Yoo, Choi, Yang, and
  Shin]{kim2015compression}
Y.-D. Kim, E.~Park, S.~Yoo, T.~Choi, L.~Yang, and D.~Shin.
\newblock Compression of deep convolutional neural networks for fast and low
  power mobile applications.
\newblock In \emph{ICLR}, 2016.

\bibitem[Kingma \& Ba(2015)Kingma and Ba]{kingma2014adam}
D.~Kingma and J.~Ba.
\newblock Adam: A method for stochastic optimization.
\newblock In \emph{ICLR}, 2015.

\bibitem[LeCun et~al.(2015)LeCun, Bengio, and Hinton]{lecun2015deep}
Y.~LeCun, Y.~Bengio, and G.~Hinton.
\newblock Deep learning.
\newblock \emph{Nature}, 521\penalty0 (7553):\penalty0 436--444, 2015.

\bibitem[Lee et~al.(2014)Lee, Sun, and Saunders]{lee2014proximal}
J.D. Lee, Y.~Sun, and M.A. Saunders.
\newblock Proximal {N}ewton-type methods for minimizing composite functions.
\newblock \emph{SIAM Journal on Optimization}, 24\penalty0 (3):\penalty0
  1420--1443, 2014.

\bibitem[Li \& Liu(2016)Li and Liu]{li2016ternary}
F.~Li and B.~Liu.
\newblock Ternary weight networks.
\newblock Technical Report arXiv:1605.04711, 2016.

\bibitem[Lin et~al.(2016)Lin, Courbariaux, Memisevic, and
  Bengio]{lin2015neural}
Z.~Lin, M.~Courbariaux, R.~Memisevic, and Y.~Bengio.
\newblock Neural networks with few multiplications.
\newblock In \emph{ICLR}, 2016.

\bibitem[Martens \& Sutskever(2012)Martens and Sutskever]{martens2012training}
J.~Martens and I.~Sutskever.
\newblock Training deep and recurrent networks with {H}essian-free
  optimization.
\newblock In \emph{Neural Networks: Tricks of the trade}, pp.\  479--535.
  Springer, 2012.

\bibitem[Novikov et~al.(2015)Novikov, Podoprikhin, Osokin, and
  Vetrov]{novikov2015tensorizing}
A.~Novikov, D.~Podoprikhin, A.~Osokin, and D.P. Vetrov.
\newblock Tensorizing neural networks.
\newblock In \emph{NIPS}, pp.\  442--450, 2015.

\bibitem[Pascanu \& Bengio(2014)Pascanu and Bengio]{pascanu2013revisiting}
R.~Pascanu and Y.~Bengio.
\newblock Revisiting natural gradient for deep networks.
\newblock In \emph{ICLR}, 2014.

\bibitem[Pascanu et~al.(2013)Pascanu, Mikolov, and
  Bengio]{pascanu2013difficulty}
R.~Pascanu, T.~Mikolov, and Y.~Bengio.
\newblock On the difficulty of training recurrent neural networks.
\newblock In \emph{ICLR}, pp.\  1310--1318, 2013.

\bibitem[Rakotomamonjy et~al.(2016)Rakotomamonjy, Flamary, and
  Gasso]{rakotomamonjy2016dc}
A.~Rakotomamonjy, R.~Flamary, and G.~Gasso.
\newblock {DC} proximal {N}ewton for nonconvex optimization problems.
\newblock \emph{IEEE Transactions on Neural Networks and Learning Systems},
  27\penalty0 (3):\penalty0 636--647, 2016.

\bibitem[Rastegari et~al.(2016)Rastegari, Ordonez, Redmon, and
  Farhadi]{rastegari2016xnor}
M.~Rastegari, V.~Ordonez, J.~Redmon, and A.~Farhadi.
\newblock {XNOR-Net}: {ImageNet} classification using binary convolutional
  neural networks.
\newblock In \emph{ECCV}, 2016.

\bibitem[{Theano Development Team}(2016)]{2016arXiv160502688short}
{Theano Development Team}.
\newblock {Theano: A {Python} framework for fast computation of mathematical
  expressions}.
\newblock \emph{arXiv e-prints}, abs/1605.02688, May 2016.
\newblock URL \url{http://arxiv.org/abs/1605.02688}.

\bibitem[Tieleman \& Hinton(2012)Tieleman and Hinton]{tieleman2012lecture}
T.~Tieleman and G.~Hinton.
\newblock Lecture 6.5-rmsprop: Divide the gradient by a running average of its
  recent magnitude, 2012.

\bibitem[Yuille \& Rangarajan(2002)Yuille and Rangarajan]{yuille2002concave}
A.L. Yuille and A.~Rangarajan.
\newblock The concave-convex procedure ({CCCP}).
\newblock \emph{NIPS}, 2:\penalty0 1033--1040, 2002.

\bibitem[Zeiler(2012)]{zeiler2012adadelta}
M.D. Zeiler.
\newblock {ADADELTA}: An adaptive learning rate method.
\newblock Technical Report arXiv:1212.5701, 2012.

\bibitem[Zhou et~al.(2016)Zhou, Ni, Zhou, Wen, Wu, and Zou]{zhou2016dorefa}
S.~Zhou, Z.~Ni, X.~Zhou, H.~Wen, Y.~Wu, and Y.~Zou.
\newblock {DoReFa-Net}: Training low bitwidth convolutional neural networks
  with low bitwidth gradients.
\newblock Technical Report arXiv:1606.06160, 2016.

\end{thebibliography}
\bibliographystyle{iclr2017_conference}

\newpage
\appendix


\section{Proof of Proposition~\ref{prop:opt}}

Denote $\|\x\|_{\bm Q}^2 = \x^\top {\bm Q} \x$,
	\begin{eqnarray*}
		\lefteqn{\nabla \ell(\hw^{t-1})^\top (\hw^t - \hw^{t-1}) + \frac{1}{2}(\hw^t - \hw^{t-1})^\top \D^{t-1} (\hw^t - \hw^{t-1})}
		\notag \\
		& = &  \frac{1}{2} \sum_{l=1}^{L}  \|\hw_l^t - (\hw_l^{t-1} -
		\nabla_l \ell(\hw^{t-1}) \oslash \d^{t-1}_l)\|_{\D_l^{t-1}}^2 + c_1
		\notag \\
		& = & \frac{1}{2} \sum_{l=1}^{L} \|\hw_l^t - \w_l^t\|_{{\D_l^{t-1}}}^2 + c_1
		\\
		& = & \frac{1}{2} \sum_{l=1}^{L} \sum_{i=1}^{n_l}  [\d_l^{t-1}]_i (\alpha_l^t [\b_l^t]_i
		- [\w_l^t]_i)^2 + c_1,
		\notag
	\end{eqnarray*}	
	where $c_1 = -\frac{1}{2} \|\nabla_l \ell(\hw^{t-1})
	\oslash \d^{t-1}_l\|_{\D_l^{t-1}} ^2 $ is independent of $\alpha_l^t$ and $\b_l^t$.
 Since  $\alpha_l^t > 0, \d_l^t \succ \bm{0},\forall l=1,2, \dots, L$, we have
$\b_l^t = \text{sign}(\w_l^t)$. Moreover,
\begin{eqnarray*}
\frac{1}{2} \sum_{l=1}^{L} \sum_{i=1}^{n_l}  [\d_l^{t-1}]_i (\alpha_l^t [\b_l^t]_i
- [\w_l^t]_i)^2 + c_1
		& = & \frac{1}{2} \sum_{l=1}^{L} \sum_{i=1}^{n_l}  [\d_l^{t-1}]_i (\alpha_l^t
		- |[\w_l^t]_i|)^2 + c_1\\
		& = & \sum_{l=1}^{L} \frac{1}{2} \|\d_l^{t-1}\|_1 (\alpha_l^t)^2 - \|\d_l^{t-1} \odot \w_l^t\|_1 \alpha_l^t + c_2, \\
\end{eqnarray*}
where $c_2 = c_1 + \frac{1}{2} \|\d_l^{t-1} \odot \w_l^t \odot \w_l^t\|_1$.  Thus, the
optimal $\alpha^t_l$ is $\frac{\|\d^{t-1}_l \odot \w_l^t\|_1}{\|\d^{t-1}_l\|_1}$.


\section{Proof of Theorem~\ref{thm:conv}}


Let $\bm{\alpha} = [\alpha_1^t \dots, \alpha_L^t]^\top$, and 
denote the objective 
in 
\eqref{eq:obj} by
$F(\hw, \bm{\alpha})$.
As $\hw^t$ is the minimizer in (\ref{eq: obj_proximal}), we have
\begin{equation}
\ell(\hw^{t-1}) + \nabla \ell(\hw^{t-1})^\top (\hw^t - \hw^{t-1}) +
\frac{1}{2}(\hw^t - \hw^{t-1})^\top \D^{t-1} (\hw^t - \hw^{t-1})
\leq \ell(\hw^{t-1}). \label{eq:p1}
\end{equation}
From Assumption A1, we have
\begin{equation}
\ell(\hw^t) \leq  \ell(\hw^{t-1}) +  \nabla \ell(\hw^{t-1})^\top (\hw^{t} - \hw^{t-1} )+ \frac{\beta}{2}\NM{\hw^t - \hw^{t-1}}{2}^2. \label{eq:p2}
\end{equation}
Using (\ref{eq:p1}) and (\ref{eq:p2}), we obtain
\begin{eqnarray*}
\ell(\hw^t) &\leq& \ell(\hw^{t-1}) - \frac{1}{2} 
(\hw^t - \hw^{t-1})^\top (\D^{t-1} - \beta \bm{I}) (\hw^t - \hw^{t-1})
\nonumber \\ 
& \leq &  \ell(\hw^{t-1}) -
\frac{\min_{k,l} ([d_l^{t-1}]_k - \beta)}{2}\NM{\hw^t - \hw^{t-1}}{2}^2.
\end{eqnarray*}
Let $c_3 = \min_{k,l,t} ([d_l^{t-1}]_k - \beta) > 0$.
Then,
\begin{align}
\ell(\hw^t) \le \ell(\hw^{t-1}) -
\frac{c_3}{2} \NM{\hw^t - \hw^{t-1}}{2}^2.
\label{eq:temp1}
\end{align}
From Assumption A2, 
$\ell$ is bounded from below.  Together with the fact that $\{\ell(\hw^t)\}$ is monotonically decreasing from \eqref{eq:temp1},
the sequence $\{\ell(\hw^t)\}$ converges, thus the sequence $\{F(\hw^t, \al^t)\}$ also converges. 

\section{Proof of Proposition~\ref{pr:singular}}
	Let the singulars values of $\W$ be $\lambda_1(\W) \geq \lambda_2(\W) \geq \cdots \geq
	\lambda_m(\W)$. 
	\[
	\lambda_1^2(\W) \geq \frac{1}{m} \sum_{i=1}^{m} \lambda_i^2(\W) =
	\frac{1}{m}\|\W\|_F^2 = \frac{1}{m}m n = n.
		\]
	Thus,
	$\lambda_1(\W) \geq \sqrt{n}$.

\end{document}